\pdfoutput=1 % ensure pdflatex (for arXiv)

\documentclass[11pt,a4paper]{article}
\usepackage{syntaxfest/syntaxfest2021}

\usepackage{nert}
\usepackage{times}
\usepackage{latexsym}

\usepackage{tikz-dependency}

\usepackage{microtype}
\usepackage{needspace}

\renewcommand{\longversion}[1]{#1}
\renewcommand{\nonanonversion}[1]{#1}
\renewcommand{\anonversion}[1]{\unskip}
\renewcommand{\finalversion}[1]{#1}
\newcommand{\futurework}[1]{}    % postponed unresolved issues

\nonanonversion{\syntaxfestfinalcopy} 
\newcommand{\deprel}[1]{\textbf{\texttt{#1}}}
\newcommand{\upos}[1]{\textsc{#1}}

\title{Mischievous Nominal Constructions in Universal Dependencies}

\author{Nathan Schneider \quad Amir Zeldes \\ %\thanks{~~\emldisplay{nathan.schneider@georgetown.edu}} \\
  Georgetown University \\
  \{\emldisplay{nathan.schneider@georgetown.edu}{nathan.schneider}, \emldisplay{amir.zeldes@georgetown.edu}{amir.zeldes}\}\texttt{@georgetown.edu} \\
}
\date{}

\begin{document}
\maketitle

\begin{abstract}
While the highly multilingual Universal Dependencies (UD) project provides extensive guidelines for clausal structure as well as structure within \emph{canonical} nominal phrases, a standard treatment is lacking for many ``mischievous'' nominal phenomena that break the mold.
As a result, numerous inconsistencies within and across corpora can be found, even in languages with extensive UD treebanking work, such as English.
This paper surveys the kinds of mischievous nominal expressions attested in English UD corpora and proposes solutions primarily with English in mind, but which may offer paths to solutions for a variety of UD languages.
\emph{An abridged version omitting \cref{sec:npmod,sec:nummeas} has been published: \citep{mncs-udw}.}
\end{abstract}

\section{Introduction}

\finalversion{
\blfootnote{
    %
    % for review submission
    %
    %\hspace{-0.65cm}  % space normally used by the marker
    % Place licence statement here for the camera-ready version. See
    % Section~\ref{licence} of the instructions for preparing a
    % manuscript.
    %
    % % final paper: en-uk version 
    %
    % \hspace{-0.65cm}  % space normally used by the marker
    % This work is licensed under a Creative Commons 
    % Attribution 4.0 International Licence.
    % Licence details:
    % \url{http://creativecommons.org/licenses/by/4.0/}.
    % 
    % % final paper: en-us version 
    %
    \hspace{-0.65cm}  % space normally used by the marker
    This work is licensed under a Creative Commons 
    Attribution 4.0 International License.
    License details:
    \url{http://creativecommons.org/licenses/by/4.0/}.
}
}

% NP exhibiting well-behaved, unproblematic relations:

% \pex{just all the/my/Sam's 6 little Labrador Retriever puppies from home (that are) playing fetch}

% det:predet (all)
% det (the)
% amod (little)
% compound (Labrador <-compound- Retriever <-compound- puppies)
% nmod (from home)
% nmod:poss (my, Sam’s)
% advmod (just)
% nummod (6)
% acl or acl:relcl

% + coordination and copulas, of course

% Our goals are:
% a) A better solution for what we call equative nominal modifiers, which are similar to appositives but do not necessarily meet all the criteria for appositives.
% b) Clearer guidelines around named entities and measurement expressions of various kinds. This includes introducing a new relation with semantic headedness criteria for dates and replacing \deprel{nmod:npmod} (sort of a grab-bag category covering rates, extents, floating reflexives, ...) with something better.
% c) Guidelines for symbols and technical notation when used in a nonsyntactic fashion.
% d) A treatment of participial and phrasal modifiers in compounds.
% \\ \az{I def agree we need guidelines for c+d. For a. I'd rather stick them under some existing relation since they're marginal. rethinking nmod:npmod is a good idea IMO, though note other non-eng TBs use it too..}

Universal Dependencies \citep[UD;][]{nivre-16,nivre-20,de_marneffe-21} is a framework describing morphology and dependency syntax cross-linguistically. It establishes common labels and structural constraints for annotating data, comparing languages, and training and evaluating parsers.

This paper, intended for readers familiar with UD (specifically, Basic Dependencies in version~2), addresses what we see as a significant shortcoming of the current guidelines: ``mischievous'' nominal structure---roughly, constructions that form noun phrases beyond the canonical components of determiner or possessive, adjective modifier, noun compound modifier, head noun or pronoun, modifier PP, and modifier clause. Many of these are productive but narrow constructions forming multiword names, dates, measurements, and compound-like structures. 

Such expressions often buck ordinary restrictions on NP structure:
\Citet{kahane-17}, for instance, note that ``most languages have particular constructions for named entities such as dates or titles\dots. These subsystems are in some sense `regular irregularities', that is, productive unusual constructions.'' 
In other words, names and dates often do not fit the mold of other noun phrases, though as we will show below, the issues they raise pop up in other environments too. For many of these mischievous constructions, the existing UD syntactic relations are inadequate, or inadequately described, and corpora are widely inconsistent as a result---in some cases within a single treebank or between treebanks in the same language.

Many of the issues presented below have been discussed at length within the UD community but without any definitive resolution.\anonversion{\footnote{Other contributors to background discussions of these issues will be acknowledged in the final version of the paper.}} Our goal is to consolidate the discussion and argue for a coherent approach (or set of alternatives) based on careful analyses of English constructions across a range of text types.\footnote{Some short examples in this paper come from introspection, while longer examples and statistics are taken from the English Web Treebank \citep[UD\_English-EWT;][]{silveira-14}, and UD\_English-GUM \citep{Zeldes2017b} or UD\_English-GUMReddit \citep{BehzadZeldes2020}, which together cover a broad spectrum of spoken and written genres and writing styles.} To minimize added complexity to the UD scheme, our proposals are conservative, focused on clarifying boundaries between existing labels and in some cases proposing new subtypes (which, though language-specific, may be adapted to other languages).  While we refrain from proposing new universal relations that would force extensive editing across languages to maintain validity,  we welcome feedback on related phenomena in other languages. Although our analysis is focused on English, we believe that similar reasoning applies to a range of other languages which cannot be adequately examined here due to space reasons; we hope that guideline discussions in those languages will benefit from the analyses below.\shortversion{\footnote{An extended version of this paper \citep{mncs-arxiv} contains additional recommendations regarding numbers and adverbial NPs, omitted here due to space limitations.}}

\section{Name Descriptors}\label{sec:desc}

We turn first to proper names, especially names of persons, and the constructions by which a speaker can elaborate on a nominal referring expression.

\begin{multicols}{2}
\ex. \a. I met Gaspard Ulliel.
    \b. I met Gaspard Ulliel, the French actor. \label{ex:appos1}
    \c. I met the French actor, Gaspard Ulliel. \label{ex:appos2}
    \z.
    
\ex. \a.\label{ex:appel-emb} I met French actor Mr.~Gaspard Ulliel.
    \b. *I met French actor. \label{ex:missing-name}
    \c. *I met the Mr.~Gaspard Ulliel.
    \z.
    
\end{multicols}

How are these handled in UD?
The \deprel{flat} relation comes into play for open-class expressions with no clear syntactic head, 
canonically including personal names like \pex{Gaspard Ulliel}.
A flat structure, by convention, is represented in UD by designating the first word as the head of each of the subsequent words, which attach to it as \deprel{flat} (a ``bouquet'' or ``fountain'' analysis).

The trouble is that referring expressions may contain descriptors beyond personals. Following the Cambridge Grammar of the English Language  \citep[CGEL;][]{cgel}, we distinguish two types of pre-name descriptors in English:
An \textbf{appellation} is a title that would be used to formally address somebody by social status (e.g.~occupation or gender), such as \pex{\textbf{Mr.} Obama} or \pex{\textbf{President} Obama}.
An \textbf{embellishment}\footnote{Also called ``false title'', described here as a kind of apposition: \url{https://en.wikipedia.org/wiki/False_title}} is a bare nominal phrase preceding the name (and appellation if there is one) describing the referent with category information like \pex{actor}, \pex{French actor}, or \pex{surprise winner of the Kentucky Derby}.\footnote{An anonymous reviewer has commented on the difficulty of applying the bare nominal diagnostic in languages with different determiner systems, such as Slavic languages, Chinese, or Japanese. We fully acknowledge that equivalent constructions may look quite different in those languages, but also believe that the problems analyzed here are both substantial enough in English to merit a more detailed treatment, and common enough in other languages that the discussion is likely relevant beyond English.} 
The embellished name may have an inanimate referent, as in \pex{\textbf{German car maker} BMW}.\footnote{Thanks to an anonymous reviewer for this example.}
In English, embellishments are characteristic of select genres such as news.\footnote{A newscaster might say, \pex{\textbf{Surprise winner of the Kentucky Derby} American Pharaoh received a hero's welcome upon returning home today\dots}. Note the lack of an article at the beginning of the sentence.}
\Cref{ex:appel-emb} contains an embellishment and an appellation within the same referring expression. The current UD guidelines state:\footnote{\url{https://universaldependencies.org/workgroups/newdoc/two_nominals.html}}

\begin{quote}
If the two nominals participate in denoting one entity, the default relation to connect them is \deprel{flat} (which may also be used to connect other nodes that are not nominals). Typical examples are personal names: we can say that \pex{John Smith} is a special type of \pex{John} as well as a special type of \pex{Smith}, but none of the names governs the other and either of them can be omitted. In many languages this analysis extends to titles and occupations, as in English \pex{president Barack Obama}.
\end{quote}

Yet the flat analysis for embellishments and appellations yields counterintuitive results. That they are bare NPs and are omissible---whereas the personal name is not, as shown by \cref{ex:missing-name}---is strong syntactic evidence that they are modifiers. Moreover, it should be intuitively obvious that \pex{Gaspard} and \pex{Ulliel} form a coherent unit of structure---yet under the bouquet analysis for flat structures (i.e.~attaching all children to the first token), \pex{Ulliel} would have distinct heads for \pex{\uline{Gaspard} Ulliel}, \pex{French \uline{actor} Gaspard Ulliel},
and \pex{\uline{Mr.}~Gaspard Ulliel}.\footnote{Note that some embellishments and appellations contain clear internal structure (e.g., \pex{\textbf{French actor} Ulliel}---\deprel{amod}; \pex{\textbf{Secretary of State} Clinton}---\deprel{nmod}, \deprel{case}). 
This does not pose an additional problem for the \deprel{flat} analysis, however: even dependents within a flat structure may host internal modifiers, as was recently clarified in the guidelines.}

\begin{figure*}
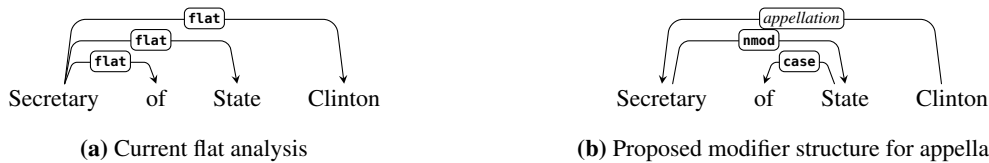
\centering
\hfill
\begin{subfigure}[b!]{0.45\textwidth}\centering
    \begin{dependency}\small
        \begin{deptext}[column sep=1.5em]
            Secretary \& of \& State \& Clinton \\
        \end{deptext}
        \depedge[edge unit distance=2ex]{1}{2}{\deprel{flat}}
        \depedge[edge unit distance=2ex]{1}{3}{\deprel{flat}}
        \depedge[edge unit distance=2ex]{1}{4}{\deprel{flat}}
    \end{dependency}
    \caption{Current flat analysis}
    \label{fig:sos-flat}
\end{subfigure}\hfill
\begin{subfigure}[b!]{0.45\textwidth}\centering
    \begin{dependency}\small
        \begin{deptext}[column sep=1.5em]
            Secretary \& of \& State \& Clinton \\
        \end{deptext}
        \depedge[edge unit distance=2ex]{1}{3}{\deprel{nmod}}
        \depedge[edge unit distance=2ex]{3}{2}{\deprel{case}}
        \depedge[edge unit distance=2ex]{4}{1}{\textit{appellation}}
    \end{dependency}
    \caption{Proposed modifier structure for appellation}
    \label{fig:sos-mod}
\end{subfigure}\hfill
\caption{An appellation with current vs.~proposed structures (several options for the relation label of the ``\textit{appellation}'' dependency discussed below).}
\end{figure*}

\futurework{\nss{His Royal Highness The Prince Philip, Duke of Edinburgh -- ``The'' is interesting. does it make it appos([The Prince], [Philip, Duke of Edinburgh])? cf. Oedipus, King of Thebes}}

Further discussion in the guidelines acknowledges treating titles as \deprel{flat} is controversial, but explains that titles do not meet typical criteria for \deprel{nmod}, \deprel{compound}, or \deprel{appos}. An \deprel{nmod} typically receives its own independent case marking (possessive or prepositional in English). \deprel{appos} is limited in UD to relations between two full NPs (or DPs, i.e.~NPs including a determiner), as in \cref{ex:appos1,ex:appos2}. And crosslinguistically, ``titles do not usually behave like compounds: in German, they are not joined to the following words, as compounds are normally joined in German, and they appear at the beginning of names in both German and Hebrew, even though German compounds are head last and Hebrew compounds are head first.''\footnote{\url{https://universaldependencies.org/u/dep/flat.html\#some-further-notes-on-relations-for-names}}
%As we will see below, the guidelines for those relations are somewhat underdeveloped, leaving many issues unclear.

Nevertheless, we suggest that appellations and embellishments be removed from the flat analysis. Exactly how this could be achieved is considered below.

\subsection{A Relation for Titles?}\label{sec:title}

A narrow solution would be to group appellations and embellishments under the category of \textbf{titles}. As these constructions are frequent and distinctive, a subtype called \deprel{:title} might be appropriate, and subtyping could alleviate the concern that none of the existing top-level deprels is a perfect fit. 
Alternatively, a new top-level relation could be introduced.
We thus begin by considering the following options:
\begin{itemize}
    \item \deprel{title}, a new top-level relation
    \item \deprel{compound:title} %. Rationale: the key syntactic characteristic of  \deprel{compound} dependents of nouns is that they are bare attributive modifiers lacking a separate article. 
    \item \deprel{appos:title}
    \item \deprel{nmod:title}
    \item \deprel{nmod:desc}, a broader subtype, meant to cover additional mischievous nominals
    %\item \deprel{bnmod:title}. Rationale: currently, \deprel{nmod} and \deprel{appos} are designed for full NPs. The \deprel{bnmod} relation would target bare nominal modifiers that do not fit \deprel{compound}.\footnote{As new top-level relations would arguably necessitate a major version update (i.e.~UDv3), \deprel{dep:title} could serve as a stopgap in the meantime.}
    
\end{itemize}

\paragraph{A new top-level relation?}
A new top-level (universal) relation, \deprel{title}, presupposes that honorific titles, at least, occur widely across languages and may have idiosyncratic syntax. %\longversion{\footnote{As new top-level relations would arguably necessitate a major version update (i.e.~UDv3), \deprel{dep:title} could serve as a stopgap in the meantime.}}
However, it seems possible that in some languages titles  might have `normal' syntax, and would not need such a top-level relation at all. Even for languages with conspicuous title syntax, UD relations aim to be as compact as possible; adding major labels is not done lightly, and would require waiting for UDv3, not to mention imposing costs on many treebank maintainers and requiring updates to existing tools. We therefore prefer subtyping an existing relation.

\paragraph{Problems with \deprel{compound:title}.}
In English, \deprel{compound} dependent nouns too are bare (lack a determiner of their own), similar to appellations and embellishments, suggesting a subtype \deprel{compound:title}. In fact, there is prior art in UD: Finnish UD documents the label \deprel{compound:nn} for appellations.\footnote{\url{https://universaldependencies.org/docs/fi/overview/specific-syntax.html\#appositions-and-appellation-modifiers}}

However, there are important differences that suggest compound nominals (at least in English) and titles are two different beasts.
While the definition of \deprel{compound} is quite vague, its applicability to modifiers of nouns is clearest in determinative compounds, either where both the head and modifier are part of a multiword proper name like \pex{Washington Post}; or where the head denotes a kind (usually, a noun that could be made either definite or indefinite) which is restricted by the modifier, e.g.~\pex{cake flavors}. Often such non-name combinations could be paraphrased with a possessive or prepositional construction if used literally (\pex{flavors of cake}); and often compounds behave like complex words and may become lexicalized as idiomatic multiword expressions.
By contrast, appellations and embellishments of proper name heads nonrestrictively add information about an entity and might be paraphrased with ``who is'' or an appositive (\pex{French actor Gaspard Ulliel $\rightarrow$ Gaspard Ulliel, the French actor / Gaspard Ulliel, who is a French actor}).

Morphosyntactic evidence also weighs against the compound analysis: English compound modifiers are very rarely plural, even when denoting multiple items---whereas appellations, embellishments, and appositives agree in number with their referent:

\ex. 
    \a.\label{ex:appel-pl} \textbf{Presidents} Obama and Biden [appellation]
    \b.\label{ex:embel-pl} French \textbf{actors} Ulliel and Marceau\footnote{It is unclear whether non-coordinated names referring to multiple individuals could license plural embellishments via semantic number agreement: \pex{An argument broke out between married actors Brad and Angelina / ?married actors Brangelina / ?British comedians Monty Python}.} [embellishment]
    \c. Sam and Isaac, my \textbf{brothers} [\deprel{appos}]
    %\c. my brothers Sam and Isaac [\deprel{appos}: see \cref{sec:appos}]
    %\d. the years 1776 and 1789 [\deprel{appos}: see \cref{sec:appos}]
    \e. *\textbf{eggs} carton(s)
    \z.

Cartons of/for multiple eggs are \pex{egg cartons}, stripping the plural ending from the compound modifier.\footnote{For exceptional pluralized modifiers in Germanic compounds see also \citet{Fuhrhop1996}.} If appellations and embellishments were special cases of the English compound construction we would expect them to resist pluralization as well, but this is not the case \cref{ex:appel-pl,ex:embel-pl}. 

%Finally from a user oriented and typological perspective, it seems likely that calling modifiable embellishments and simple appellations such as `Ms.' compound modifiers will appear outlandish to many linguists interested in using UD treebanks to collect data on compounding, further reducing the attractiveness of this classification.

\paragraph{Problems with \deprel{appos:title}.}

Part of the practical motivation for the \deprel{appos} relation is to express a semantic notion of equivalence between referring expressions, such that an information extraction system could strip out supplementary information when matching names against entities in a knowledge base. Thus \pex{French actor Gaspard Ulliel, my hero since childhood, won an Oscar} could be simplified to \pex{Gaspard Ulliel won an Oscar} by removing \deprel{appos} and \deprel{appos:title} dependents. From an argument structure perspective, \deprel{appos} is characterized by not adding participants to valency frames, i.e.~\pex{Gaspard Ulliel} and \pex{my hero since childhood} both instantiate the subject of \pex{won}.

On the other hand, \deprel{appos} is already rather complicated (spelled out in detail below, \cref{sec:appos}). 
While embellishments are sometimes categorized as appositions, there is a lack of universal agreement that appellations and embellishments qualify as appositive modifiers; other sources \citep[e.g.,][p.~77]{ruppenhofer-16} view the name rather than the embellishment as the appositive phrase.

\paragraph{Intermediate Proposal: a subtype of \deprel{nmod}.}
The rationale here is that \deprel{nmod} is the most general relation for nominals modifying other nominals. (It already has subtypes, including \deprel{nmod:poss} for possessive modifiers and \deprel{nmod:tmod} for temporal modifiers.)
In English, plain \deprel{nmod} dependents have case marking or prepositions, but the subtyping can signal a morphosyntactically exceptional construction, as is already the case with prepositionless \deprel{nmod:tmod}.

If we target only titles, then \deprel{nmod:title} is the least objectionable solution narrowly tailored for embellishments and appellations, given that (a)~\deprel{nmod} already has other subtypes, (b)~this would avoid confusion with dominant uses of \deprel{compound} and \deprel{appos}, and (c)~implementing a new universal relation across treebanks would be onerous, but treebanks are allowed flexibility to diverge and innovate with subtypes. On the other hand, there are a number of other `mischievous' adnominal constructions requiring a solution, which suggests that a subtype focusing only on titles may be too narrow, motivating a more general name fitting other types of descriptive modifiers, for which we will propose a new relation (called \deprel{nmod:desc}).

\subsection{Other Special Types of Nominal Modification}\label{sec:bare}

The above discussion is limited to appellations and embellishments that precede a name. But other, less frequent constructions bear some resemblance to these: % despite being \emph{post}modifiers:

\ex.\label{ex:postname} Post-name bare nominal modifiers:
    \a.\label{ex:scion} 11-year-old \uline{Draco}, \textbf{scion of the Malfoy family}, was sorted into Slytherin.
    \b.\label{ex:thebes} \uline{Oedipus}, \textbf{King of Thebes}
    \z.

\ex.\label{ex:pronnoun} First or second person pronoun plus noun:\footnote{Elsewhere the pronouns are analyzed as determinatives \citep[p.~374]{cgel}. We are concerned that it would be counterintuitive to extend \deprel{det} (and perhaps the \upos{det} tag) to include such specialized uses of \w{we}, \w{us}, and \w{you}, which are familiar to annotators as pronouns.
Nevertheless, \Citet{hohn-21}, discussing this construction at length for English and other languages, advocates the \deprel{det} solution over \deprel{appos} or \deprel{nmod}.}
    \a.\label{ex:pilots} \uline{We} \textbf{pilots} deserve a pay raise.
    \b. \uline{You} \textbf{guys} deserve a pay raise.\footnote{The expression \pex{you guys} has been conventionalized in some dialects as a gender-neutral second person plural.}
    \z.

In \cref{ex:postname,ex:pronnoun}, the bolded nominal phrase can be omitted while its head (underlined) cannot.
\Cref{ex:scion} can be considered a post-head embellishment, and \cref{ex:thebes} a post-head appellation.
The construction seen in \cref{ex:pronnoun}, headed by a pronoun, is a cousin of the pre-head embellishment, as shown by the third person paraphrase of \cref{ex:pilots}: \pex{\textbf{pilots} \uline{Earhart and Lindbergh}}. A broad relation \deprel{nmod:desc} for the special cases seen above as well as appellations and embellishments would separate them from the \deprel{appos}, \deprel{compound}, and \deprel{flat} cases while covering sufficient ground to merit its inclusion. %\nss{What about calling it \deprel{nmod:elab} for ``elaborator'', since ``descriptor'' is already used by CGEL with a narrower meaning?}

% We do not suggest \deprel{compound:bare} because \emph{all} compound modifiers are bare.
% If one of the \cref{sec:title} strategies is adopted, then the \deprel{compound} relation could be applied to \cref{ex:postname,ex:refl,ex:pronnoun}, perhaps subtyped as \deprel{compound:post} as the phrase order is exceptional relative to other compounds.
% It is worth noting, however, that the semantics is also unlike typical compounds: the modifier can be understood as in an identity relationship with the head, rather than subcategorizing it.

% \paragraph{Noun + Reflexive Pronoun.}
% English reflexive pronouns have, apart from their main function in clause structure, a function of emphasis when postmodifying an NP, e.g.~\pex{\textbf{the king himself} made the decision}.
% This postmodifying reflexive should be attached to the head noun as \deprel{neqmod:refl}.
% \az{I recognize this is a construction, and certainly an interesting one, but a label just for this would be too sparse IMO. }

% PRONOUN + NOUN: https://github.com/amir-zeldes/gum/issues/71
% \paragraph{Pronoun + Noun.}
% Finally, consider the construction combining a first or second person pronoun with a bare nominal category referent: 

\subsection{More on Appositives}\label{sec:appos}

A classic example of an appositive appears in \cref{ex:sam}. The appositive phrase, \textit{my brother}, is a nonrestrictive full NP descriptor of \textit{Sam}.
It is syntactically omissible, and could in fact replace its head as they share the same referent. A similar phenomenon appears in \cref{ex:sam2}, where an indefinite NP ascribes a property to \textit{Sam}:

\ex.\label{ex:sam} \uline{Sam}, \textbf{my brother}, is very tall.

\ex.\label{ex:sam2} \uline{Sam}, \textbf{a musician}, is very tall.

The current definition of the \deprel{appos} relation establishes the following criteria:

\ex.\label{ex:appos-criteria} An appositive (\deprel{appos}) must be
    \a.\label{ex:app-full} a full NP
    \b.\label{ex:app-rev} modifying an NP in a reversible fashion (modulo punctuation)
    \c.\label{ex:app-right} to the right
    \d.\label{ex:app-immed} with no intervening words.\footnote{An exception to this constraint is already found in languages with so-called Wackernagel particles, such as Classical Greek or Coptic, which appear in the second position in the sentence and can interrupt any phrase or dependency; see \citet{ZeldesAbrams2018}.}
    \z.

While appositive phrases are often separated by commas or parentheses, this is not a strict requirement, and of course spoken language has no commas. We understand the definition to also include:

\ex.\label{ex:app-categorizing} 
    \a. my \uline{brother} \textbf{Sam}
    \b. the \uline{color} \textbf{purple}
    \c. the \uline{word} ``\textbf{terrorist}''
    \d. the \uline{play} \textbf{\emph{Much Ado About Nothing}}
    \z.

Cases resembling appositives in some but not all of the above respects require clarification. The bare modifiers discussed above are sometimes considered appositives, but UD excludes them with criterion \cref{ex:app-full}.
\cref{ex:amy} satisfies criteria \cref{ex:app-full,ex:app-right} but not \cref{ex:app-rev,ex:app-immed}, whereas \cref{ex:leader} satisfies \cref{ex:app-full,ex:app-rev,ex:app-immed} but fails \cref{ex:app-right}:

\ex.\label{ex:amy} ``Maybe she really does just need a little space\dots,'' \uline{Amy} said, \textbf{ever the optimist}.\footnote{\emph{The Body in the Casket: A Faith Fairchild Mystery}, Katherine Hall Page, 2017}

\ex.\label{ex:leader} \textbf{A new Pakistani leader}, \uline{he} is intent on instituting reforms.

There seem to be two ways forward:
\begin{itemize}
    \item Relax \deprel{appos} criteria either in general or in a subtype. In particular, relaxing \cref{ex:app-rev,ex:app-right,ex:app-immed} would allow \deprel{appos} to cover \cref{ex:amy,ex:leader}.
    This would contrast with \deprel{nmod:desc} suggested above, which covers bare nominal modifiers.
    %Compare the possible new subtypes noted in \cref{sec:title} (\deprel{appos:title}) and \cref{sec:bare} (\deprel{appos:desc}), which would be exempt from criterion \cref{ex:app-full}.
    
    \item Maintain the \deprel{appos} criteria in \cref{ex:appos-criteria}, and classify examples such as \cref{ex:amy,ex:leader} as \deprel{dislocated}.
    These constructions are not quite classic dislocation constructions,\footnote{The preferatory appositive in \cref{ex:leader}---which features a description followed by a definite NP, and would be perfectly at home in a newspaper---is not to be confused with hanging topic left-dislocation with a pronoun referring back to the dislocated element, as might be uttered in conversation: \textit{My dad, he is always running late.}} but they could be treated as if removed from their normal apposition location.
\end{itemize}

In the interest of maintaining the status quo for appositions, we favor the latter solution and recommend using \deprel{dislocated}.

%\longversion{\az{Can we say we prefer the latter? I know I do :)}\nss{I'm fine with dislocated, but what is the justification? conservativism with the definition of appos, or something else?}}

\section{Further Issues with Names}
%%%% tracking issue: https://github.com/UniversalDependencies/docs/issues/757

\begin{table*}\centering\small\setlength{\tabcolsep}{2pt}
\begin{tabular}{@{}l@{ }lclllcp{6.5em}@{}}
 & & \textbf{head} & \multicolumn{1}{c}{\textbf{modifier optional?}} & \multicolumn{1}{c}{\textbf{invertible?}} & \multicolumn{1}{c}{\textbf{agreement?}} & \textbf{type}  & \textbf{relation} \\
\midrule
\cref{ex:appel-emb} & actor Ulliel & R & Ulliel & *Ulliel, actor  & actors Ulliel and Marceau & name (head) & \deprel{$\leftarrow$nmod:desc} \\
\cref{sec:desc} & President Obama & R & Obama & *Obama, President  & Presidents Obama and Biden & name & \deprel{$\leftarrow$nmod:desc} \\
\cref{sec:names-analyzable} & Church Street & R & *Street / the street & *Street, Church  & Church and River Streets & name & \deprel{$\leftarrow$compound} \\[5pt]
\cref{ex:enttype} & Lake Michigan & L & *Lake / the lake & *Michigan, Lake & Lakes Michigan and Ontario & name & \deprel{compound$\rightarrow$} \\
\cref{ex:type-num} & Figure 4 & L & *Figure / the figure & *4, Figure & Figures 4 and 5 & name w/~num & \deprel{nummod:name$\rightarrow$}? \deprel{compound$\rightarrow$}? \deprel{nmod:desc$\rightarrow$}? \\
\cref{ex:name-num} & Firefox 58.0 & L & Firefox & *58.0, Firefox & *Firefoxes 58.0 and 59.0 & name w/~num & \deprel{nummod:name$\rightarrow$}? \deprel{flat}? \deprel{nmod:desc$\rightarrow$}? \\
\cref{sec:addr} & London, UK & L & London & *UK, London & *Londons, UK and Ontario & name & \deprel{nmod\shortversion{:npmod}\longversion{:adv}$\rightarrow$} \\
 & Joe Biden & -- & (flat) & (flat) & *Joe and Jill Bidens & name & \deprel{flat} \\
\cref{ex:sam} & my brother Sam & L & my brother & Sam, my brother & my brothers Sam and John & name (mod) & \deprel{appos$\rightarrow$} \\
\end{tabular}
\caption{Constructions involving names and their syntactic properties.}
\label{tab:names}
\end{table*}

\subsection{Syntactically analyzable proper names}\label{sec:names-analyzable}

Several other aspects of the syntax of names need to be addressed. 
The syntactic properties of many of the constructions at issue 
are summarized in \cref{tab:names}. We begin by underscoring UD's policy of analyzing the internal structure of names with ordinary syntax where possible, regardless of the semantic status of the name.
%UD trees are meant to describe syntactic structure, not lexical semantics. 
%As such, multiword proper names are analyzed as following ordinary syntactic patterns where possible.
For example, \pex{Church Street} is analyzed with \deprel{compound}; and  \pex{New York City} consists of an %\az{I think in EWT New York is still flat, which GUM copies}\nss{Nope it's amod in EWT. ``New'' still looks like an adjective modifier even if its meaning is somewhat opaque}
adjective which modifies a noun (\deprel{amod}), which in turn modifies another noun (\deprel{compound}).\footnote{Previously, POS tags in the English treebanks followed Penn Treebank tags and treated all content words within a proper name as \upos{propn}, but this was changed in v2.8; \upos{propn} is now limited to nouns.}

%(This exception does not apply to proper adjectives---``words that are derived from names but are adjectives rather than nouns'', e.g.~\pex{\textbf{European} trade}---which are tagged as \upos{adj} and modify nouns as \deprel{amod}.\footnote{\url{https://universaldependencies.org/u/pos/ADJ.html}})

\subsection{Cardinal directions}

Cardinal direction modifiers of nouns (\pex{north}, \pex{northeast}, etc.)\ are annotated inconsistently in English UD corpora. Based on the tagging tradition of LDC corpora, these should be treated as nouns unless they bear overt adjectival morphology (\pex{northern}, etc.). Cardinal direction nouns premodifying nouns should therefore attach as \deprel{compound}, whether the expression is a proper name (\pex{North Carolina}) or not (\pex{north coast}). When multiple parts of a cardinal direction term are separated by a space or hyphen, they are joined with \deprel{compound}: e.g.~\pex{north east} `northeast'.

\subsection{Names beginning with an entity type}\label{sec:enttype}

Many proper names incorporate a transparent entity type.
In \pex{the Thames River}, the name is constructed as an ordinary endocentric compound, with the entity type last and serving as the head and an identifier as the modifier.\footnote{Other place names headed by an entity type and exhibiting ordinary syntax include \pex{Mirror Lake}, \pex{Ford's Theatre}, and \pex{the Dome of the Rock}.} But \pex{the River Thames} (along with the other examples in \cref{ex:enttype}) poses a problem as the order is reversed:

\ex.\label{ex:enttype} %Names beginning with the entity type:  % https://github.com/UniversalDependencies/docs/issues/501
    \a. Mount Fuji
    \b. Fort Knox
    \c. Lake Michigan
    \d.\label{ex:river-thames} the River Thames
    \z.

It can be argued that the head in \cref{ex:river-thames} is then \w{Thames}, as \w{River} can be omitted: \pex{the Thames} \citep[pp.~519--20]{cgel}.
However, this omission of the entity type could be viewed as a shortening not unlike reducing \pex{Fenway Park} to \w{Fenway} on the assumption that the speaker is able to identify the referent based on the more specific part of the name.
Such shortenings will vary in felicitousness depending on the particular name and context. 
(Plain \w{Michigan} does not refer to the same thing as \pex{Lake Michigan}.)

Note also that the name-initial entity types may be pluralized when grouping together multiple entities of the same type, 
which distinguishes them from flat structures or typical compound modifiers and suggests they may be heads:
\pex{\textbf{Lakes} Michigan and Ontario} (cf.~\pex{Mirror and Swan \textbf{Lakes}}).
This fits with the expected semantics, as noun-noun compounds tend to be headed by the superordinate category, and historically it is possible that the construction is in fact a remnant of left-headed compounding from Romance place names, possibly from Norman toponym patterns (English \pex{Mount \emph{X}}, French \pex{Mont-\emph{X}}, e.g.~\pex{Mont-Saint-Michel}).

We therefore consider the examples in \cref{ex:enttype} as inverted (left-headed) compounds.\footnote{Another analysis we considered was to treat the entity type as an \deprel{nmod:desc} modifier, giving \pex{Lake Michigan} the same structure as \pex{Dr.~Livingstone} or \pex{actor Ulliel}. But the entity types in \cref{ex:enttype} seem more central to the name than titles, and are not as freely omissible, so we are not persuaded that they are modifiers.}
The identifier can attach to the entity type as \deprel{compound} to reflect the inverted word order in these kinds of names.

%\nss{aren't compound modifiers usually omissible without adding determiners? is this a problem?} -- never mind, cf. names like Church Street

% \Cref{ex:enttype} illustrates a special class of English proper name patterns in which an entity type precedes the open-class part of the name.\footnote{Contrast \pex{Mirror Lake}, \pex{Ford's Theatre}, \pex{the Mississippi River}, and \pex{the Dome of the Rock}, all of which are headed by the entity type and follow ordinary syntactic patterns.}
% CGEL refers to these as \textbf{descriptors} \citep[pp.~519--20]{cgel}.
% These are combinations of nouns, so on the surface \deprel{compound} should apply. 
% But standard English compounds are headed by the superordinate category, whereas for these the entity type is the modifier.
% Moreover, coordinated instances may share a plural entity type (\pex{\textbf{Forts} Knox and Bragg}), recalling the plural appellations illustrated in \cref{ex:appel-pl}.
% (They are perhaps less routinely omissible than 
% personal name appellations, though: plain \pex{Knox} may or may not be apparent as an abbreviated reference to \pex{Fort Knox}.)

\subsection{Numbered entities}\label{sec:numid}
% https://github.com/UniversalDependencies/docs/issues/466 "section X"
% https://github.com/UniversalDependencies/docs/issues/654 "number X"

Numbers can also figure into names. 
They can disambiguate multiple of a series of related 
entities named by a proper noun, as in \cref{ex:name-num}. These are appendages to a proper name, syntactically omissible (with a resulting broadening of meaning), and could be treated as modifiers. 
Numbers can also follow an entity type, as in \cref{ex:type-num}.

%\par\vspace{-10pt}\par

\begin{multicols}{2}
\ex.\label{ex:name-num} \a.\label{ex:version} Firefox (version)~58.0
     \b.\label{ex:richard-iii} Richard~III
     \c. \emph{Toy Story~3}
     \d. 1~Corinthians
     \e. World War II
     \z.

\columnbreak

\ex.\label{ex:type-num} \a. Figure~4
     \b. room~11b
     \c.\label{ex:range} pp.~5--10
     \d. subpart~(e)
     \e.\label{ex:item-no} item (number)~3
     \f.\label{ex:symphony} \emph{Symphony No.~5}
     \z.

\end{multicols}

%\par\vspace{-10pt}\par

The cases in \cref{ex:type-num} use the number to identify a specific instance of the type. The entity type appears first, similar to the inverted \deprel{compound} examples in \cref{sec:enttype}. It is a completely different construction from  quantity modification, the predominant application of \deprel{nummod},  as in \pex{3 items} (plural!) or \pex{3\%}. A morphosyntactic difference between the numeric modifier constructions in \cref{ex:name-num} and \cref{ex:type-num} is that only the latter exhibit agreement: \pex{page~5} (one page), \pex{pages~5--10} (multiple pages), but \pex{*Firefoxes 58.0 and 59.0}.

We see three options, each with pros and cons:
\begin{itemize}
    \item The morphosyntactic difference notwithstanding, treat \cref{ex:name-num} and \cref{ex:type-num} as essentially the same construction, with a new relation such as \deprel{nummod:name} (consistent with the fact that the superordinate category \deprel{nummod} is currently applied to numeric modifiers generally).\footnote{The choice of subtype parallels \deprel{flat:name}---an optional subtype not currently implemented in English corpora, though it is used for a number of corpora in other languages. The \href{https://universaldependencies.org/u/dep/flat-name.html}{\deprel{flat:name} guidelines}
currently include \pex{Formula~\textbf{1}} as an example; this would become \deprel{nummod:name} in this option.} 
    Advantages are that \cref{ex:name-num} and \cref{ex:type-num} look very similar, and numbers are a salient property for annotators or corpus users to notice when selecting the appropriate relation. However, adding a subtype for a relatively narrow and infrequent phenomenon is questionable, and some cases are not numeric (\pex{Level B}).
    
    \item Treat \cref{ex:name-num} and \cref{ex:type-num} as instances of more general constructions. 
    The construction in \Cref{ex:type-num} can be considered an inverted compound like \pex{Lake Michigan} (\cref{sec:enttype}).
    Flat structures could apply to the names in \cref{ex:name-num} as this construction is less morphologically transparent. 
    This would avoid a new subtype but also may be seen as splitting hairs based on a subtle morphosyntactic criterion.
    
    \item A third option is to adopt \deprel{nmod:desc} for the constructions in \cref{ex:name-num} and \cref{ex:type-num}. 
    This would essentially restrict the definition of \deprel{compound} to substantive lexical material excluding numbering designators; \deprel{nmod:desc} would broadly cover miscellaneous modifiers associated with names that do not fit the more conventional constructions. 
    This solution eclipses the similarity between \pex{Lake Michigan} (which would remain \deprel{compound}) and \pex{Figure~4}, but it perhaps avoids a counterintuitively broad application of \deprel{compound}.  It also means that the scope of \deprel{nmod:desc} is a bit broader, including not just modifiers that are secondary to the main part of a name, but also modifiers that are essential to it (just \pex{Figure} is not a name, whereas \pex{Ulliel} is).
\end{itemize}

%A numbering scheme---possibly arbitrary, possibly associated with a meaningful scale or temporal ordering---may be used to distinguish instances of a category on a formal or ad hoc basis:

% Semantically, the number helps identify a particular instance; unlike ordinary \deprel{nummod}, it is not a quantity.
% And superficially, the examples in \cref{ex:type-num} and \cref{ex:name-num} appear similar, with the number appended (usually after) the content noun.
% But delving deeper, we can observe that the singular entity type noun cannot stand on its own in \cref{ex:type-num}, 
% whereas the number can be omitted in \cref{ex:name-num}.
% And the entity type noun in \cref{ex:type-num} pluralizes to agree with groups of numeric identifiers; contrast \pex{*Firefoxes 58.0 and 59.0}.\footnote{Some names may be pluralized; a web search, for example, indicates that \pex{Elizabeths I and II} is attested though \pex{Elizabeth I and II} is preferred.}
% Thus the \cref{ex:type-num} examples exhibit similarities to descriptors in \cref{sec:desc}, and should be analyzed as \deprel{nmod:desc}, with the number as head.
% For examples in \cref{ex:name-num}, where the number is a modifier in the name, we suggest \deprel{nummod:name} \footnote{The choice of subtype parallels \deprel{flat:name}---an optional subtype not currently implemented in English corpora, though it is used for a number of corpora in other languages. The \href{https://universaldependencies.org/u/dep/flat-name.html}{\deprel{flat:name} guidelines}
% currently include \pex{Formula~\textbf{1}} as an example; this would become \deprel{nummod:name} under our proposal.}.

\Cref{ex:version,ex:item-no,ex:symphony} illustrate a construction in which a word like \w{number} or \w{version} may precede a number to clarify that it is an identifier rather than a quantity.
In modern usage this would generally remain singular even if referring to multiple items (\pex{items number 3 and 4}), so we analyze \w{number} as a \deprel{compound} modifier by default, and \deprel{nmod:desc} only if plural (\pex{items numbers 3 and 4}). %\az{I think it's nm:dsc items->number AND nm:dsc number->3}\nss{You don't like the agreement test for :desc? Is there a syntactic reason to exclude compound if it's singular?} \az{I thought about it and don't feel strongly; I guess only if you read it as ellipsis -- `items numbers 3 and 4' would be an embellishment in any case, and you could read `items number 3 and 4' as ellipsis for `items number 3 and (number) 4', so the singular version is actually two coordinate embellishments. But I'm totally fine if you prefer compound here.}
``?'' is provided as a stand-in for the relation between the entity type and the number given the above uncertainty:\footnote{Confirming native speaker intuitions, a search of COCA \citep{coca} reveals that the plural is much less frequent than the singular in the pattern \pex{\textsc{n.pl} number(s) \textsc{num} and \textsc{num}}, with the exception of the abbreviated spelling, where \w{nos.}\ is more prevalent in this context than \w{no.}\ (the abbreviations seem to be especially conventional in proper names like \pex{Symphony No.~5}).}

% Certain kinds of names instead license \w{version} in this position \cref{ex:version}.
% These can be treated like number-modifying appellations.

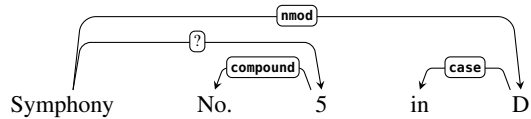
\begin{center}
\begin{dependency}\small
    \begin{deptext}[column sep=3em]
        Symphony \& No. \& 5 \& in \& D \\
    \end{deptext}
    \depedge[edge unit distance=2.2ex]{1}{3}{?}
    \depedge[edge unit distance=2ex]{3}{2}{\deprel{compound}}
    \depedge[edge unit distance=1.7ex]{1}{5}{\deprel{nmod}}
    \depedge[edge unit distance=2ex]{5}{4}{\deprel{case}}
\end{dependency}
\end{center}

For hyphenated numeric ranges \cref{ex:range}, the prevailing policy in UD corpora has been to analyze the second part like a prepositional phrase \pex{to 10}, thus an \deprel{nmod} of \w{5}. 
\nonanonversion{One of the authors takes the view}\anonversion{Another view is}
that a coordination analysis would be more natural. 
In any event, \w{5} attaches to \w{pp.}\ as a modifier.
%\az{I would try to make a decision here, and I'm willing to accept either compound or nmod:desc}\nss{I don't feel prepared to make a decision. I see this as something to seek input on in the true spirit of a workshop paper. :)}

%\deprel{nummod:id} attaches \w{pp.}\ to \w{5}.

% Note that the \pex{number X} pattern in \cref{ex:item-no} behaves differently from \cref{ex:number-three}:

% \ex. \a.\label{ex:number-three} This is my \textbf{number~3} priority. (`3rd priority')
%     \b. *This is my 3 priority.
%     \c. *This is my number priority.
%     \z.

% This is a special construction which on balance seems like a type of compounding that turns a cardinal number into the equivalent of an ordinal.
% Thus we suggest a nested compound analysis:

% \begin{center}
% \begin{dependency}\small
%     \begin{deptext}[column sep=3em]
%         my \& number \& 3 \& priority \\
%     \end{deptext}
%     \depedge[edge unit distance=1.7ex]{4}{1}{\deprel{nmod:poss}}
%     \depedge[edge unit distance=2ex]{3}{2}{\deprel{compound}}
%     \depedge[edge unit distance=2ex]{4}{3}{\deprel{compound}}
% \end{dependency}
% \end{center}

%\nss{``3rd'' would be amod + NumType=Ord. ``number 3'' is essentially an analytic ordinal. but maybe this doesn't get expressed syntactically and it's compound(priority, 3) + whatever we do above for ``number''}

\subsection{Business and personal name suffixes}\label{sec:persname}

Adjective-expanding suffixes like \w{Inc.}\ (``incorporated'') in \pex{Apple Inc.}\ 
should attach as \deprel{amod}.
Nominal suffix designations that do not head the name, e.g. \w{LLC} (``limited liability corporation''), 
should attach as \deprel{nmod:desc}. For personal names, the suffix type \pex{III} in \cref{ex:richard-iii} is addressed above. Generational name suffixes that do not use numerals, like \pex{Richard \textbf{Jr.}}\ and \pex{Richard \textbf{the Third}}, are treated as postmodifying \deprel{amod}. Other abbreviated name suffixes that would expand to nominal expressions, such as professional or honorary designations (\w{MD}, \w{O.B.E.}), attach as \deprel{nmod:desc}.

\subsection{Nicknames and parenthetical descriptors}

A nickname that takes the form of a full NP appended to a name, e.g.~\pex{Richard \textbf{the Lionheart}}, 
can be attached as \deprel{appos}. The same goes for works of art featuring a formulaic name followed by a nickname: \pex{Symphony No.~5 ``\textbf{Fate}''}. Parenthetical descriptions following a name that are not alternate references to the entity should be treated as \deprel{parataxis}: \pex{Pierre Vinken, \textbf{61}, said\dots}; \pex{Vinken, \textbf{61 years old}, said\dots}; 
\pex{The Chicago Manual of Style, \textbf{17th edition}};
\pex{Biden~\textbf{(D)} said\dots} (but \pex{Biden, \textbf{a Democrat}, said\dots} would be \deprel{appos}).

\subsection{Addresses}\label{sec:addr}

A street address like \pex{221b Baker St.}\ is headed by \pex{St.}, with \pex{Baker} attaching as \deprel{compound}, and \pex{221b} per the policy on numbered entities (\cref{sec:numid}). Frequently, place descriptions specify a locale-NP postmodifier without a connective word besides punctuation. Examples: \pex{London, \textbf{UK}}; \pex{University of Wisconsin–\textbf{Madison}}; \pex{CSI: \textbf{Miami}}.
%The new relation \deprel{nmod:desc} comes in handy here:
These should be considered adverbial NPs, 
\longversion{with the new relation \deprel{nmod:adv} introduced in \cref{sec:npmod}.}\shortversion{which arguably should fall under the \deprel{nmod:npmod} relation.\footnote{Currently, corpora sometimes use \deprel{nmod:npmod} and sometimes use \deprel{appos}, which is not appropriate as the two parts of the location are not interchangeable.
Space does not permit full discussion of \deprel{nmod:npmod} here \citep[but see][\S 6]{mncs-arxiv}.
}}

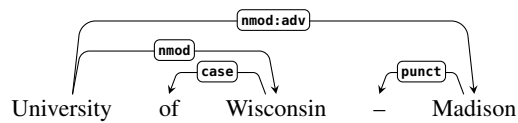
\longversion{
\begin{center}
\begin{dependency}\small
    \begin{deptext}[column sep=1.5em]
        University \& of \& Wisconsin \& – \& Madison \\
    \end{deptext}
    \depedge[edge unit distance=2ex]{1}{3}{\deprel{nmod}}
    \depedge[edge unit distance=2ex]{3}{2}{\deprel{case}}
    \depedge[edge unit distance=2ex]{5}{4}{\deprel{punct}}
    \depedge[edge unit distance=1.7ex]{1}{5}{\deprel{nmod\shortversion{:npmod}\longversion{:adv}}}
\end{dependency}
\end{center}
}

%\nss{todo: how to connect the distinct parts of an address: name - street - city - postal code? list?}

Multiple tokens of a single phone number should be joined with \deprel{flat} (this is the practice in the GUM corpus; EWT currently favors \deprel{nummod}). Separate pieces of metadata that are juxtaposed in an extralinguistic fashion (e.g., name, street address, city, postal code) should be treated as items of a list---successive items should attach to the first as \deprel{list}.

% \textbf{Washington}: The White House announced that \dots [dateline] --- obl:addr if frequent enough?

%\nss{miscellaneous formats, e.g. automobile Year-Make-Model (1994 Honda Accord): nmod:tmod and compound}\nss{actually I'd just use compound}

\section{Phrasal Attributive Modifiers}\label{sec:compounds}

% https://github.com/UniversalDependencies/docs/issues/753

\begin{figure*}
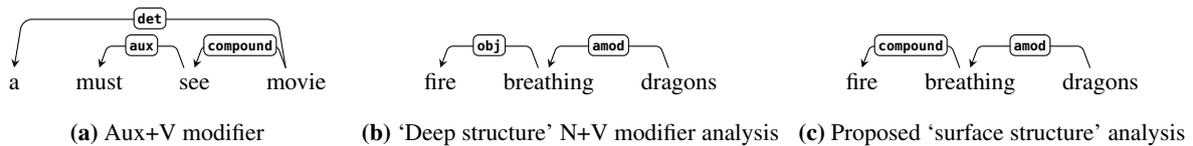
\centering
\hfill
\begin{subfigure}[b]{0.3\textwidth}\centering
    \begin{dependency}\small
        \begin{deptext}[column sep=1.9em]
            a \& must \& see \& movie \\
        \end{deptext}
        \depedge[edge unit distance=1.5ex]{4}{1}{\deprel{det}}
        \depedge[edge unit distance=2ex]{3}{2}{\deprel{aux}}
        \depedge[edge unit distance=2ex]{4}{3}{\deprel{compound}}
    \end{dependency}
    \caption{Aux+V modifier}
    \label{fig:must-see}
\end{subfigure}\hfill
\begin{subfigure}[b]{0.35\textwidth}\centering
    \begin{dependency}\small
        \begin{deptext}[column sep=1.5em]
            fire \& breathing \& dragons \\
        \end{deptext}
        \depedge[edge unit distance=2ex]{2}{1}{\deprel{obj}}
        \depedge[edge unit distance=2ex]{3}{2}{\deprel{amod}}
    \end{dependency}
    \caption{`Deep structure' N+V modifier analysis}
    \label{fig:fire-breathing-deep}
\end{subfigure}\hfill
\begin{subfigure}[b]{0.33\textwidth}\centering
    \begin{dependency}\small
        \begin{deptext}[column sep=1.5em]
            fire \& breathing \& dragons \\
        \end{deptext}
        \depedge[edge unit distance=2ex]{2}{1}{\deprel{compound}}
        \depedge[edge unit distance=2ex]{3}{2}{\deprel{amod}}
    \end{dependency}
    \caption{Proposed `surface structure' analysis}
    \label{fig:fire-breathing-surface}
\end{subfigure}\hfill
\caption{Phrasal attributive modifiers (hyphen tokens omitted for brevity).}
\label{fig:phrasal}
\end{figure*}

In English, the attributive modifier position before the noun head in a noun phrase is not limited to adjectives/adjective phrases (\pex{very easy to use}) and nominals.
It also accommodates phrases like:
%\par\vspace{-10pt}\par

\needspace{5em}
\ex.\label{ex:phrasal} 
\begin{multicols}{2}
    \a.\label{ex:phr-adjnoun} a \textbf{high-quality} \uline{product}
    \b.\label{ex:phr-pp} a \textbf{by-the-book} \uline{strategy}
    \c.\label{ex:phr-vp1} a \textbf{fly-by-night} \uline{operation}
    \d.\label{ex:phr-vp2} a \textbf{have-your-cake-and-eat-it-too} \uline{plan} % kill-your-darlings
    \e.\label{ex:phr-imp1} a \textbf{come-to-Jesus}, \textbf{do-or-die} \uline{moment}
    \f.\label{ex:phr-imp2} a stern \textbf{don't-mess-with-me} \uline{look}
    \f.\label{ex:must-see} a \textbf{must-see} \uline{movie}
    \f.\label{ex:fire-breathing} \textbf{fire-breathing} \uline{dragons}
    \f.\label{ex:church-going} the \textbf{Bible-thumping}, \textbf{church-going} \uline{faithful}
    \f.\label{ex:so-called} many \textbf{so-called} \uline{libertarians}
    \f.\label{ex:nuclear-free} a \textbf{cost-effective}, \textbf{nuclear-free} \uline{future}
    \z.
\end{multicols}

Assuming that the hyphenated expressions are tokenized as separate words, UD annotators are confronted with two issues: how to analyze these phrases internally, and which dependency relation to use for the modification of the external noun.

Some of the hyphenated expressions in \cref{ex:phrasal} are clearly lexicalized; others are productive combinations.
Expressions of this type might loosely be described as `compounds', in the sense that the joining of multiple content words into one lexical item is the morphological process of compounding.
Should the hyphenated parts thus be joined together with \deprel{compound} across the board? 
We are hesitant to establish this policy because it would overload an already very broad relation label.
Centrally, in noun phrases, \deprel{compound} describes modification of a noun by another noun.
If it applies to the examples in \cref{ex:phrasal}, it would be for attachment to the underlined noun, not the internal structure of the hyphenated expression.

Another consideration is that the internal structure of the hyphenated phrases is \emph{largely} regular: phrasal modifiers of nouns can be structured as modified nouns \cref{ex:phr-adjnoun}, PPs \cref{ex:phr-pp}, VPs \cref{ex:phr-vp1,ex:phr-vp2}, imperative sentences \cref{ex:phr-imp1,ex:phr-imp2}, and verb clusters \cref{ex:must-see}.
These structures are transparent, and just as UD policy analyzes regular internal structures in proper names like \pex{University of Wisconsin}, we advocate recognizing internal structure here.

Yet synthetic or argument structure compounds such as \pex{fire-breathing}, \pex{Bible-thumping}, and \pex{church-going} \cref{ex:fire-breathing,ex:church-going} invert the normal clausal order. Neither \w{fire} nor \w{Bible} nor \w{church} is the subject in the clausal paraphrase: \w{fire} is the direct object in \pex{breathing fire}; the paraphrase of \pex{Bible-thumping} would require reordering and adding a determiner or plural for the direct object; and the paraphrase of \pex{church-going} would require a preposition: \pex{going \textbf{to} church}. Meanwhile, \pex{so-called} \cref{ex:so-called} lacks any obvious paraphrase as a clause.
We take these anomalies in word order and morphosyntax as clear evidence that left-headed `deep structure' VP material is being grafted onto a right-headed compound in the `surface structure'. 
As Basic UD aims to represent surface syntax, we join these expressions as \deprel{compound},
as shown for \pex{fire-breathing} in \cref{fig:fire-breathing-surface} (vs.~\cref{fig:fire-breathing-deep}). %\az{I think this is a totally traditional and uncontroversial position BTW, Germanic deverbal synthetic compounds are very common and virtually always discussed in the literature as a subtype of normal compounds}\nss{if you think this is worth mentioning could you add a citation?}
The adjective-headed combinations in \cref{ex:nuclear-free} should also use internal \deprel{compound}, as should numeric modifier compounds like \pex{a \textbf{10-year} plan}.\footnote{Contrast \pex{10-year} (\deprel{compound}) with \pex{10 years}  (\deprel{nummod}), where the number modifier controls agreement.}

The next question is the external attachment, which is made difficult by UD's lexicalist principle that the part of speech of a word determines which relations it can participate in. 
Consider \pex{must-see} \cref{ex:must-see}, which is not a full VP, merely an auxiliary plus its head verb.
Is this to be treated as a clausal dependent---\deprel{acl}, or even \deprel{acl:relcl} (a relative clause)?
This seems dubious; note that a relative clause paraphrase would involve an embedded subject, e.g.~\pex{a movie that \textbf{one} must see}, or else a passive---\pex{a movie that must be seen}.
It is also doubtful whether \cref{ex:phr-vp1,ex:phr-vp2,ex:phr-imp1,ex:phr-imp2} should be treated as clausal modification, yielding several different dependency labels for the attributive relationship.
A simpler solution, it seems to us, is to treat attributive phrasal expressions internally headed by verbs like coerced %adjective % Why? See my comment in the Google doc - I think it's like "hit movie" or "blockbuster movie" - it doesn't grade like an adjective ("??a very must see movie"), and it works by itself as an NP head too: ("It's a must see"). I think I can even pluralize it: "All the must sees I missed this year". I went ahead and changed this but if you disagree let's talk about it again.
noun phrases,\footnote{\label{fn:extpos}\Citet{kahane-17} suggest expanding the UD notion of multiword token to include idiomatic phrasal expressions, separating their external syntactic behavior from their internal structure. This would make it convenient to represent the expression \pex{must-see} as a multiword \upos{noun} comprised internally of an \upos{aux} and a \upos{verb}. This could be indicated via a morphological feature \texttt{ExtPos=NOUN} on the internal head, \w{see}.}
 with %\deprel{amod} 
 \deprel{compound} for the external attachment, as shown in \cref{fig:must-see}.
As for PP modifiers like in \cref{ex:phr-pp}, it seems simplest to attach 
them as \deprel{compound} rather than \deprel{nmod}; on this view,
English nominal \deprel{compound} is equivalent to attributive modification by a non-possessive nominal phrase
(a hypothetical alternate name being \deprel{nmod:attr}).

To summarize, our proposed policy for phrasal attributive modifiers of nouns is:
\begin{itemize}
    \item The attributive expression is internally analyzed with regular relations to the extent possible, except where those relations defy ordinary word order or morphosyntax. \deprel{compound} is used internally for anomalous relations.
    \item In the interest of simplicity, all non-possessive attributive modifiers attach as either \deprel{compound} if internally headed by a nominal or nominalized phrase (including PPs), and \deprel{amod} etc.\  %\nss{why ``etc.''?} 
    for adjectival heads, as appropriate. %\nss{TODO:} %\az{um, actually I just prefer compound across the board. Adj as modifier in Germanic compounds is pretty rare, I'd assume VP modifiers are coerced into NPs..}\nss{well we have clear cases of lexicalized amod combinations (hot dog). ``fire-breathing'' can be a gerund and ``must-see'' can refer to a thing (these are must-sees), so in those contexts they are more nominal than adjectival. But it feels like the attributive uses are not exploiting the nominalizations---a fire-breathing dragon is more directly a dragon characterized by breathing fire, than a dragon that performs fire-breathing. And I'm not sure ``come-to-Jesus'' or ``don't-mess-with-me'' can be nominalized at all. So if we treated these as single words I think they'd be ADJ or VERB amods.}
\end{itemize}

%We agree with CGEL's suggestion \citep[p.~1660]{cgel} to treat such irregularly structured attributive phrases as compound adjectives, and thus would use \deprel{amod}: --- actually CGEL here was not talking about 'compound verbs'

%\nss{CGEL pp. 444, 1660}

\futurework{\nss{non-endocentric compounds: `the London--Glasgow express' \citep[p.~1660]{cgel}: internally, conj? flat? compound?}}

\section{Dates}\label{sec:dates}

While analytically expressed dates like \pex{the thirty-first of July} follow normal syntax 
(with \pex{thirty-first} elliptical for \pex{thirty-first day}),
there are special written formats for dates and times.
Instead of a flat structure, which would obscure the compositionality of dates,
we propose the simple principles of (a)~treating the most precise part of the expression as its head, 
and (b)~connecting the parts of the expression together with \deprel{nmod:tmod}.\footnote{We considered finer-grained relations like \deprel{nmod:month}, \deprel{nmod:year}, \deprel{nmod:era}, \deprel{nmod:ampm}, and \deprel{nmod:tz} but concluded these were too detailed for UD and should fall under the purview of information extraction.}

For example, \pex{July 31, 1980 AD} consists of a year expression (\pex{1980~AD}) and a month both modifying a date:
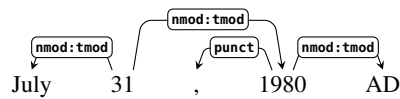
\begin{center}
\begin{dependency}\small
    \begin{deptext}[column sep=2em]
        July \& 31 \& , \& 1980 \& AD \\
    \end{deptext}
    \depedge[edge unit distance=2ex]{2}{1}{\deprel{nmod:tmod}}
    \depedge[edge unit distance=2ex]{4}{3}{\deprel{punct}}
    \depedge[edge unit distance=2.3ex]{2}{4}{\deprel{nmod:tmod}}
    \depedge[edge unit distance=2ex]{4}{5}{\deprel{nmod:tmod}}
\end{dependency}
\end{center}
Another convention puts the date before the month (\pex{31 July}).
There, too, the date would be the head.
Even when the date is written as an ordinal---\pex{July the fourth}---the month should be considered a temporal modifier because it can be omitted with sufficient context (\pex{I'll see you on the fourth}; \pex{*I'll see you on July}). 
This is in contrast to \pex{Richard the Third} (\cref{sec:persname}), where \pex{Richard} is the head.

A further practical consideration is that UD tree heads are often used to determine minimal token spans for annotations such as entity recognition, mentions in coreference resolution, and entity linking spans for Wikification \citep[associating mentioned entities with their Wikipedia entries;][]{RatinovEtAl2011}. Such minimal or `MIN' spans \citep[p.~12]{PoesioEtAl2018} are then used for training and scoring systems in `fuzzy' match scenarios. It makes intuitive sense for the day in date expressions to form the minimal span which needs to be identified, since the other tokens, i.e.~years and months, already form the minimal spans for the nested mentions of those years and months as separate entities. This use of UD-tree heads is already in place for non-UD corpora using UD parses, such as ARRAU \citep{UryupinaEtAl2019}, and in the gold standard UD English GUM for NER, coreference and Wikification \citep{LinZeldes2021}.

For time expressions we follow similar reasoning, with an example as follows:
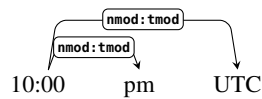
\begin{center}
\begin{dependency}\small
    \begin{deptext}[column sep=2em]
        10:00 \& pm \& UTC \\
    \end{deptext}
    \depedge[edge unit distance=2ex]{1}{2}{\deprel{nmod:tmod}}
    \depedge[edge unit distance=2.3ex]{1}{3}{\deprel{nmod:tmod}}
\end{dependency}
\end{center}
The time zone could alternately be expressed as a phrase like \pex{London time}, which we would also view as \deprel{nmod:tmod}.
If written as \pex{ten o'clock}, the token \pex{o'clock} is considered an adverb and \deprel{advmod} of \pex{ten}. This also corresponds to an etymological reading of \pex{o'clock} (< \pex{of clock}), since a univerbized prepositional phrase is equivalent to an adverb (cf.~adverbs like \pex{ashore}, formed with the Old English preposition \pex{an}, the stressed equivalent of \pex{on}).

\Citet{zeman-21} likewise proposes a standard for dates and times (considering English as well as Czech, Indonesian, and Chinese). That approach is similar, differing mainly in treating the year in a date expression as headed by the month rather than the date---\pex{1980} would be a dependent of \pex{July}, which would be a dependent of \pex{31}, in \pex{July 31, 1980}.
While semantically intuitive (smaller units of time head the next larger containing unit), it is not clear that there is any \emph{syntactic} motivation to group the month and year together. 
Although the month cannot normally be omitted while retaining the year, an expression like \pex{the 31st, 1980} is only semantically nonsensical, or at best pragmatically anomalous, but not truly ungrammatical. As evidence for this we consider the possibility of felicitous day+year expressions, such as \pex{New Year's Day 2000} (the same as 2000-01-01) or \pex{Pentecost 2022} (2022-06-05).
The year-modifies-month approach also has the disadvantage of creating nonprojectivity if the date is written between the month and the year.

\Citeauthor{zeman-21} (\S5) suggests \deprel{appos} to link a date with a day of the week, as in \pex{Wednesday, July 31}. 
We agree with this policy.
Though the day of the week conventionally comes first in English, we recognize that the order may be reversed on occasion (reversibility is a definitional criterion for \deprel{appos}, which is always left-headed). 
Moreover, this does not affect preposition choice, as \pex{on} marks days of the week as well as dates, supporting the \deprel{appos} analysis in which they are essentially interchangeable full NPs.

%\nss{existing relation, captures temporality. Most instances without case marking are on the right (``nice post \textbf{today}''; citations in GUM) but a few on the left: ``Q1 timetable'', ``1990 report''.}

\section{How prevalent are these issues?}

Some readers may wonder how common the issues raised \shortversion{in this paper}\longversion{thus far} actually are, and in particular whether their frequency merits adding relation subtypes such as \deprel{nmod:desc}. \Cref{tab:descstats} gives statistics for some types of constructions that would be covered under the umbrella of such a relation. Although the phenomena are not extremely frequent, the total token count of 373 out of 152K tokens in the UD v2.9 edition of GUM puts a putative relation covering these at rank 35 of 49 relation labels (including subtypes), between \deprel{obl:tmod} (362 tokens) and \deprel{nmod:tmod} (399), suggesting that these are not particularly rare occurrences. We also presume that depending on genre, some subtypes may become much more frequent, such as company suffixes or even personal titles---for example, the frequency of just company suffixes in EWT seems is about 2.5 per 10K tokens, compared to 0.3 per 10K tokens in GUM (other categories are harder to identify, since their annotation in EWT currently varies or is not easily distinguishable, as in the case of numbering modifiers).

\begin{table*}[h!tb]\centering\small\setlength{\tabcolsep}{5pt}
\begin{tabular}{llcc}
\textbf{construction} & \textbf{most frequent types}        & \textbf{tokens (GUM)} & \textbf{types (GUM)} \\
\midrule
title/profession      & General (15), Mr. (10), St. (8)     & 202                   & 78                   \\
numbering             & Figure (31), Method (20), Wave (10) & 162                   & 63                   \\
company               & Inc (4)                             & \hphantom{00}4                     & \hphantom{0}1                    \\
entity type           & Mount (1), Camp (1), Team (1)       & \hphantom{00}5                     & \hphantom{0}5                    \\
\midrule
\textbf{total}        &                                     & 373                   & 147                 
\end{tabular}
\caption{Frequencies of some mischievous nominal constructions in GUM.}
\label{tab:descstats}
\end{table*}

Although adding a new labeling distinction in the form of \deprel{nmod:desc} would doubtless require some manual disambiguation effort, we feel that by surveying the constructions in this paper in detail, it becomes more feasible to design high recall, automatic approaches to creating an initial updated version of UD English with a more nuanced treatment of these mischievous constructions, using UD editing libraries such as DepEdit \citep{PengZeldes2018} or Udapi \citep{PopelZabokrtskyVojtek2017}, which can then be subjected to a manual filtering pass.

\longversion{  % hide sections 6 & 7 to keep within 10 pages
\section{Adverbial NPs}\label{sec:npmod}

\begin{figure*}
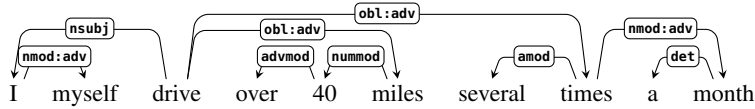
\centering\small
\begin{dependency}\small
    \begin{deptext}[column sep=1em]
        I \& myself \& drive \& over \& 40 \& miles \& several \& times \& a \& month \\
    \end{deptext}
    \depedge[edge unit distance=2ex]{1}{2}{\deprel{nmod:adv}}
    \depedge[edge unit distance=2.3ex]{3}{1}{\deprel{nsubj}}
    \depedge[edge unit distance=2ex]{5}{4}{\deprel{advmod}}
    \depedge[edge unit distance=2ex]{6}{5}{\deprel{nummod}}
    \depedge[edge unit distance=1.5ex]{3}{6}{\deprel{obl:adv}}
    \depedge[edge unit distance=2ex]{8}{7}{\deprel{amod}}
    \depedge[edge unit distance=2ex]{10}{9}{\deprel{det}}
    \depedge[edge unit distance=2.3ex]{8}{10}{\deprel{nmod:adv}}
    \depedge[edge unit distance=1.2ex]{3}{8}{\deprel{obl:adv}}
\end{dependency}
\caption{Illustration of constructions with adverbial NP modification (after renaming \deprel{:npmod} to \deprel{:adv}), and an approximator modifying a number.}
\label{fig:npmod}
\end{figure*}

We have seen a variety of difficult kinds of modification within a noun phrase. 
It is necessary to consider another kind of modification wherein an \textit{entire noun phrase} serves as the modifier.
The current UD English guidelines apply the \deprel{:npmod} subtype for these, 
resulting in two relations: \deprel{nmod:npmod} for NP modifiers within a larger NP or PP, and \deprel{obl:npmod} for others.
\Cref{fig:npmod} illustrates some of these with our proposed renaming of \deprel{:npmod} to \deprel{:adv}, as justified below.

English-specific guidelines enumerate five subcategories of \deprel{:npmod}.\footnote{The \href{https://universaldependencies.org/en/dep/nmod-npmod.html}{\deprel{nmod:npmod}} guidelines list (omitting examples):
\emph{(i)~a measure phrase, which is the relation between the head of an adjectival/adverbial or prepositional phrase and the head of a measure phrase modifying it;
(ii)~noun phrases giving an extent to a verb, which are not objects;
(iii)~financial constructions involving an adverbial, notably the following construction \emph{\$5 a share}, where the second nominal means “per share”;
(iv) floating reflexives;
and (v) certain other absolutive nominal constructions.}
(The \deprel{obl:npmod} guidelines page omits (iii).)}
But in practice, its application is both heterogeneous and inconsistent.

\subsection{EWT survey}

We examined the English Web Treebank (EWT), the largest gold-standard UD reference corpus for English \citep{ewtb,silveira-14}, 
to identify common uses of \deprel{nmod:npmod} (163~instances) and \deprel{obl:npmod} (573~instances).
The main ones are as follows:
\begin{itemize}
    \item Extent modifier, where ``extent'' is defined broadly to mean a degree, hedge, spatial distance, temporal duration, frequency, number of repetitions, or other measure such as an amount of money.
    Some of the most prevalent subcases:
    
    \ex.\label{ex:dimadj} Modifying a dimensional adjective or adverb (\textit{long}, \textit{deep}, \textit{old}):
    \textbf{6m} \uline{deep}; \textbf{a week} \uline{longer}; I am \textbf{17 years} \uline{old} (\deprel{obl:npmod}, 51~tokens)
    
    \ex. Modifying \pex{early}, \pex{earlier}, \pex{late}, \pex{later}, or \pex{sooner}:
    We arrived \textbf{10 minutes} \uline{early}; She remembers him \textbf{32 years} \uline{later} (\deprel{obl:npmod}, 35~tokens)
    
    \ex. Modifying \pex{away}, \pex{apart}, \pex{ago}, or \pex{back}: 
    \textbf{several metres} \uline{away}; \textbf{two decades} \uline{ago} (\deprel{obl:npmod}, 62~tokens)
    
    \ex. Modifying a PP or subordinate clause:
        \a. \textbf{two months} before the \uline{election}; \textbf{some miles} to the \uline{west}; \textbf{a little} out of my \uline{way} (\deprel{nmod:npmod}, 36~tokens)
        \b. \textbf{ten minutes} before they \uline{closed} (\deprel{obl:npmod}, 16~tokens\footnote{In 10 of these the head is the subordinating conjunction rather than the subordinate predicate---seemingly an error given UD's reluctance for function words to be heads.})
        \z.
    
    \ex. Degree modifier \pex{a} + \pex{lot}\slash\pex{little}\slash\pex{bit}\slash\pex{touch}\slash\pex{notch}: 
    \textbf{a lot} \uline{harder}; \uline{scaring} me \textbf{a bit} (\deprel{obl:npmod}, 103~tokens not counted above)
    
    \ex. Hedge modifier of adjective or predicate: I was \textbf{kind of} \uline{curious}; She \textbf{sort of} \uline{apologized} (\deprel{obl:npmod}, 15~tokens)
    
      %     \item The fixed-expression hedges \pex{kind of} and \pex{sort of}, where \pex{kind} and \pex{sort} are tagged as nouns.
    \item Rates (measure phrase + indefinite NP):
        \ex. With measure phrase: \uline{\$}30 \textbf{an entree}; 3 \uline{times} \textbf{a week} (\deprel{nmod:npmod}, 12~tokens)
        
        \ex. With \textit{once} or \textit{twice}:	\uline{once} \textbf{a week} (\deprel{obl:npmod}, 10~tokens)
        
    \item Adverbial reflexive pronouns: 
        \ex. you may remove the \uline{tumor} \textbf{itself} surgically (\deprel{nmod:npmod}, 30~tokens)
    
        \ex. I \uline{trained} her \textbf{myself} (\deprel{obl:npmod}, 14~tokens)
    
    \item \pex{way}-NPs: \uline{Drove} \textbf{all the way}; I didn't expect to \uline{react} \textbf{that way}; \textbf{Either way}\dots (\deprel{obl:npmod}, 25~tokens)
    \item Adverbial idioms internally headed by a noun: I \uline{fell} \textbf{head over heels} (\deprel{obl:npmod}, 10~noun+preposition+noun tokens)
\end{itemize}

In all of these, the NP can be said to have an adverbial function.\footnote{Except that \cref{ex:dimadj} covers instances like \pex{a \textbf{3-4 month} \uline{old} kitten} and \pex{my \textbf{4 year} \uline{old}}, which should be \deprel{compound} (\cref{sec:compounds}).} 
But the \deprel{advmod} relation is defined to narrowly cover lexical adverb modifiers, 
and so is not available for caseless adverbial NPs, and \deprel{nmod} is subtyped instead.
We suggest \deprel{:adv} as a more coherent and less confusing subtype.\footnote{An alternative to subtyping \deprel{nmod} and \deprel{obl} would be to rely on the external POS (\texttt{ExtPos}) morphological feature already in use by some UD treebanks (\cref{fn:extpos}). Applying \texttt{ExtPos=ADV} to a (pro)noun would convey that it heads a phrase which acts externally like an adverb, 
making it a valid \deprel{advmod} dependent. This would reduce the number of subtyped dependency relations by moving information into the morphological features. %, with potentially greater expressivity for the treatment of multiword expressions generally.
}

Additional constructions unearthed in our \deprel{:npmod} survey are:
\begin{itemize}
    \item Time zone postmodifiers in dates, better treated as \deprel{nmod:tmod} as described in \cref{sec:dates}.
    \item Compounds between a noun and an adjective, participle, gerund, or other verb. These do not actually involve a full NP as modifier, and are better analyzed with \deprel{compound} as described in \cref{sec:compounds}.
    \item NPs expressing supplementary or parenthetical information (typically set off by punctuation) that should be \deprel{parataxis}, e.g.: 
        \ex. \uline{Ronald} Joseph Crawford, \textbf{42,} of Hamilton
    
        \ex. the mythical perception that \uline{war} \textbf{-{}- especially nuclear war -{}-} was around the corner
    
        \ex. One minister reportedly \uline{handed} out 100 dollar \textquotesingle gifts\textquotesingle{} to journalists attending a press conference for Allawi, \textbf{a practice that brings back bad memories to many Iraqis}.
    
    \item Paratactic or non-syntactic juxtaposition of information, which should be \deprel{parataxis} or \deprel{list}:
        \ex. \uline{Copyright} 2005 \textbf{Houston Chronicle}
        
        \ex. \uline{did} a very Professional job very quick, \textbf{no fuss}
        
        \ex. For Curr \uline{LME} LME (Spot) 01Mar01 \textbf{JPY}/USD
        
\end{itemize}
All told, perhaps 200--300 of the EWT instances should be changed to something other than \deprel{:adv} under these guidelines. 
Thus, \deprel{:adv} is significantly narrower than \deprel{:npmod} as currently applied.

%\finalversion{A full breakdown of constructions from the survey appears in \nss{appendix}.}
% https://docs.google.com/spreadsheets/d/19fz9bAJqYFH1CRk0k575m5OeP38WiSJK713ZVyVoyus/edit#gid=0

\subsection{Relationship to \deprel{:tmod}}

Prepositions are optional or impossible for many noun-headed temporal modifiers of events (\pex{He arrived (on) \textbf{Thursday}}; \pex{He worked here (for) \textbf{an hour}}; \pex{He arrived \textbf{today}}).
Such caseless temporal modifier NPs receive distinct subtypes \deprel{obl:tmod} and \deprel{nmod:tmod}. %(which are considered special cases of the \deprel{:npmod} relations).
The current documentation, however, is not entirely clear on the scope of the \deprel{:tmod} subtype---should it apply in more general constructions where one of the modifiers happens to be temporal, e.g.~compounds (\pex{a \textbf{2018}/research paper}), rates (\pex{\$15 an \textbf{hour}/inch}), and geographic distances (\pex{10 \textbf{minutes}/miles away})?

We believe that as UD's goal is to represent syntax, rather than semantics,
\deprel{:tmod} is best limited to \emph{adverbial} modifiers that are temporal.
This includes points in time, frequencies, and durations of events, 
whether the temporal modifier is a specific date expression or measurement with units, or a vague description like \pex{several times} or \pex{a while}. 
It also includes modifiers of temporal adverbs such as \pex{late(r)} and \pex{ago}.
%But it excludes modifiers of temporal adverbs such as \pex{late(r)} and \pex{ago} (which ordinarily feature temporal modifiers for semantic, not syntactic, reasons).
Thus:

\ex. \a. my \uline{schedule} \textbf{yesterday}: \deprel{nmod:tmod}
    \b. \textbf{This week} I \uline{work} \textbf{3 times}: \deprel{obl:tmod}
    \c. a \textbf{2018/research} \uline{paper}: \deprel{compound}
    \d. \a. \uline{\$15} \textbf{an inch}: \deprel{nmod:adv}
        \b. \uline{\$15} \textbf{an hour}: \deprel{nmod:tmod}
        \z.
    \e. \a. \textbf{10 miles} \uline{away}: \deprel{obl:adv}
        \b. \textbf{10 minutes} \uline{ago/away}: \deprel{obl:tmod}
        \z.
    \f. \a. \textbf{a lot} \uline{taller}: \deprel{obl:adv}
        \b. \textbf{a lot} \uline{later}: \deprel{obl:tmod}
    \z.

%\az{I think the temporal :adv cases are too confusing; it's just counter-intuitive to me that `10 minutes ago' is not tmod. I would just say tmod is :adv whenever the thing is temporal (but still only do it with things that don't have an ADP)}\nss{i was on the fence here. you think it should be, first decide whether it's an adverbial NP, and if it's temporal then :tmod otherwise :adv?}\az{yes, that seems the easiest thing to explain, basically I think that the :tmod thing is really a 100\% semantic, legacy thing}
} % end of adverbials/:npmod section

\longversion{
\section{Numbers and Measurements}\label{sec:nummeas}

\subsection{Multiword Numbers}

% https://github.com/UniversalDependencies/docs/issues/198

Special, language-specific patterns govern the linguistic expression of numbers.
As a ``torture test'' consider \pex{835,101.596}, which could be read as 
\pex{eight hundred thirty-five thousand, one hundred and one point five nine six}.
What is the correct parse?
Current English UD guidelines specify \deprel{compound} for \pex{\textbf{four} \uline{thousand}} and \pex{\textbf{3.2}~\uline{billion}},\footnote{\url{https://universaldependencies.org/en/dep/compound.html}}\footnote{In the SUD framework \citep[intended to be convertible to and from UD;][]{sud}, spelled-out numbers are treated as flat structures: \url{https://surfacesyntacticud.github.io/guidelines/u/particular_phenomena/compounds/}} 
indicating that \deprel{nummod} should be reserved for the attachment of the full number expression to its external head.
But for longer numbers, the guidelines do not discuss whether to nest complex substructures or how to treat decimal points. No precedent is apparent as numbers with multiple spelled-out elements (rather than numerals) are rare in the English UD corpora.\footnote{\Citet{smith-99}, however, offers an analysis of number names in HPSG.}

\begin{figure*}
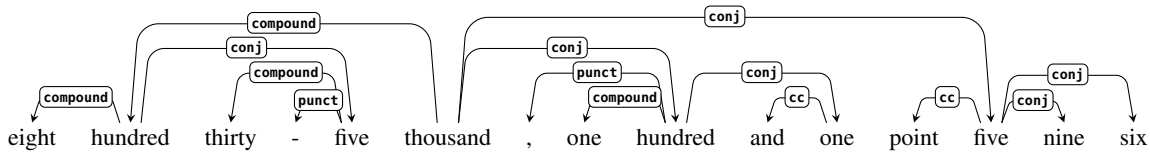
\centering
\begin{dependency}\small
    \begin{deptext}[column sep=1em]
        eight \& hundred \& thirty \& - \& five \& thousand \& , \& one \& hundred \& and \& one \& point \& five \& nine \& six \\
%       1        2          3         4    5       6           7    8      9          10     11     12       13
    \end{deptext}
    \depedge[edge unit distance=2.3ex]{2}{1}{\deprel{compound}}
    \depedge[edge unit distance=2.3ex]{5}{3}{\deprel{compound}}
    \depedge[edge unit distance=2ex]{5}{4}{\deprel{punct}}
    \depedge[edge unit distance=2.3ex]{2}{5}{\deprel{conj}}
    \depedge[edge unit distance=2.3ex]{6}{2}{\deprel{compound}}
    \depedge[edge unit distance=2.3ex]{9}{7}{\deprel{punct}}
    \depedge[edge unit distance=2.3ex]{9}{8}{\deprel{compound}}
    \depedge[edge unit distance=2.3ex]{6}{9}{\deprel{conj}}
    \depedge[edge unit distance=2.3ex]{11}{10}{\deprel{cc}}
    \depedge[edge unit distance=2.3ex]{9}{11}{\deprel{conj}}
    \depedge[edge unit distance=2.3ex]{13}{12}{\deprel{cc}}
    \depedge[edge unit distance=1.4ex]{6}{13}{\deprel{conj}}
    \depedge[edge unit distance=2ex]{13}{14}{\deprel{conj}}
    \depedge[edge unit distance=2.2ex]{13}{15}{\deprel{conj}}
\end{dependency}
\caption{A spelled-out version of the number 835,101.596. If serving as a quantity modifying a noun, the head (\pex{thousand}) would attach to the noun as \deprel{nummod}.}
\label{fig:number}
\end{figure*}

In \cref{fig:number} we suggest a tree for our example. 
It maintains \deprel{compound} for multiplicative combinations like \pex{one hundred}, 
and also uses it for \emph{tens}-\emph{units} hyphenations like \pex{thirty-five}.
Other additive or sequential-digit combinations, whether expressed with a coordinating conjunction or via juxtaposition, 
are analyzed with \deprel{conj}.
In this way the tree can be constructed such that subexpressions are apparent.
The POS category of \w{point} for the decimal separator is not entirely clear; we treat it as a coordinating conjunction.

\subsection{Approximators}

Approximators are modifiers that alter the bounds of a quantity or measure, 
e.g., \pex{\textbf{more than} \uline{3} books}, \pex{\textbf{under} a \uline{minute}}, \pex{\textbf{at least} \uline{once}}, \pex{a price of \textbf{about} \uline{\$}10}. They attach as \deprel{advmod} to the closest plausible quantity modified, which could be a number\longversion{ (as in \cref{fig:npmod})}, dollar sign, or the head of an indefinite measure phrase like \pex{a \uline{year}}.\footnote{In their approximator usages, \pex{more than}, \pex{less than}, and \pex{up to} are treated as adverbial \deprel{fixed} expressions, while \pex{at least} and \pex{at most} are treated as PPs, per current guidelines (\url{https://universaldependencies.org/en/dep/fixed.html}).}

\futurework{\subsection{Ranges}

\nss{hyphens; X to Y; between X and Y; from X to Y}}

\subsection{Units}

% \nss{8' 10" etc.}
% \nss{8 degrees ENE}
% \nss{80 F; 80 degrees Fahrenheit}

It is a general policy in UD that symbols which would be pronounced as a word are tokenized and treated syntactically like that word---e.g., \pex{\$10} read as ``ten dollars'' receives the \deprel{nummod} relation with \pex{\$} as the head.
Similarly, the notation \pex{5\textquotesingle 11"}---meaning a height of 5~feet, 11~inches---is split into four tokens. The additive juxtaposition of the two measurements should be \deprel{conj} following the principle in the previous section.

Terms like \pex{degrees Fahrenheit} should be analyzed with a rightward-pointing \deprel{compound} relation, similar to the old-fashioned word order in \pex{the brothers Grimm}, i.e.~`the Grimm brothers'.
The expression \pex{110~F} (with no separate token for the degrees) should be analyzed with \deprel{nummod}, as if the \pex{F} stands for  ``Fahrenheit-degrees''.
} % end of numbers and measurements section

% \section{Emphatic reflexives}

% \nss{are these focus markers?}

% \ex.\label{ex:refl} Post-nominal reflexive pronoun adding emphasis:\nss{Amir: this should be :npmod}
%     \a. The \uline{king} \textbf{himself} made the decision.
%     \b. \uline{I} \textbf{myself} am ready to leave.
%     \z.

% \section{Symbols and Technical Notation}

% \nss{some written sentences contain textual semiotic systems other than language. how to connect, e.g., mathematical formulas; code fragments; musical chord progressions; moves in chess notation: \deprel{list}?}

\section{Conclusion}

Above we have reviewed many constructions involving names, values,\shortversion{ and} compounds\longversion{, and adverbial noun phrases}
that have pointed to blind spots in the current guidelines for the \deprel{nmod:*}, \deprel{compound}, \deprel{flat}, \deprel{appos}, and \deprel{nummod} relations.
We have laid out several options for improving the treatment of these constructions via clearer and more principled guidelines.
The proposed improvements are of a surgical nature, minimizing disruption to other UD conventions 
(no new universal relations are proposed, for instance).
We are cognizant that considerable effort may be required to fully revise existing UD treebanks, 
but note that treebanks are already inconsistent; clearer guidance can only help. 
Subtypes remain officially optional---it is not necessary for a treebank to distinguish subtypes of \deprel{nmod} to be compliant with the UD standard.

We invite feedback on these proposals from the UD community, particularly with regard to other languages. 
We are aware that treebanking efforts in other languages have encountered some of the same issues, 
but we have not systematically investigated our proposed solutions beyond English.

\nonanonversion{
\section*{Acknowledgments}

Though the synthesis of problems, argumentation, and recommendations presented above are new,
many of the specific challenges and alternatives arose from background discussions on GitHub about the UD guidelines. 
These discussions took place over several years in more than a dozen discussion threads involving (at least):
Aryaman Arora, 
Colin Batchelor, 
Flavio Massimiliano Cecchini, 
Xinying Chen, 
Çağrı Çöltekin, 
Sylvain Kahane, 
Chris Manning, 
Bohdan Moskalevskyi,
Joakim Nivre,
Martin Popel, 
Alexandre Rademaker, 
Livy Real, 
Jack Rueter,
Sebastian Schuster, 
Francis Tyers, 
Jonathan North Washington, 
and Dan Zeman. 
Emily M.~Bender offered a suggestion of related work.
We are grateful to all who provided insights.
}

\bibliography{mncsUD}

\begin{thebibliography}{24}
\expandafter\ifx\csname natexlab\endcsname\relax\def\natexlab#1{#1}\fi

\bibitem[{Behzad and Zeldes(2020)}]{BehzadZeldes2020}
Shabnam Behzad and Amir Zeldes. 2020.
\newblock \href {https://aclanthology.org/2020.wac-1.7} {A cross-genre ensemble
  approach to robust {R}eddit part of speech tagging}.
\newblock In \emph{Proc. of the 12th Web as Corpus Workshop}, pages 50--56,
  Marseille, France.

\bibitem[{Bies et~al.(2012)Bies, Mott, Warner, and Kulick}]{ewtb}
Ann Bies, Justin Mott, Colin Warner, and Seth Kulick. 2012.
\newblock \href {https://catalog.ldc.upenn.edu/LDC2012T13} {English {W}eb
  {T}reebank}.
\newblock Technical Report {LDC2012T13}, Linguistic Data Consortium,
  Philadelphia, {PA}.

\bibitem[{Davies(2010)}]{coca}
Mark Davies. 2010.
\newblock \href {http://llc.oxfordjournals.org/content/25/4/447} {The {C}orpus
  of {C}ontemporary {A}merican {E}nglish as the first reliable monitor corpus
  of {E}nglish}.
\newblock \emph{Literary and Linguistic Computing}, 25(4):447--464.

\bibitem[{Fuhrhop(1996)}]{Fuhrhop1996}
Nanna Fuhrhop. 1996.
\newblock Fugenelemente.
\newblock In Ewald Lang and Gisela Zifonun, editors, \emph{Deutsch -
  typologisch}, pages 525--550. de Gruyter, Berlin.

\bibitem[{Gerdes et~al.(2018)Gerdes, Guillaume, Kahane, and Perrier}]{sud}
Kim Gerdes, Bruno Guillaume, Sylvain Kahane, and Guy Perrier. 2018.
\newblock \href {https://www.aclweb.org/anthology/W18-6008} {{SUD} or
  {Surface-Syntactic} {U}niversal {D}ependencies: an annotation scheme
  near-isomorphic to {UD}}.
\newblock In \emph{Proc. of the Second Workshop on Universal Dependencies
  ({UDW} 2018)}, pages 66--74, Brussels, Belgium.

\bibitem[{H\"{o}hn(2021)}]{hohn-21}
Georg F.~K. H\"{o}hn. 2021.
\newblock \href {https://aclweb.org/anthology/2021.udw-1.6} {Towards a
  consistent annotation of nominal person in {U}niversal {D}ependencies}.
\newblock In \emph{Proc. of the Fifth Workshop on Universal Dependencies (UDW,
  SyntaxFest 2021)}, pages 75--83, Sofia, Bulgaria.

\bibitem[{Huddleston and Pullum(2002)}]{cgel}
Rodney Huddleston and Geoffrey~K. Pullum, editors. 2002.
\newblock \href
  {https://archive.org/details/TheCambridgeGrammarOfTheEnglishLanguage}
  {\emph{The {C}ambridge {G}rammar of the {E}nglish {L}anguage}}.
\newblock Cambridge University Press, Cambridge, {UK}.

\bibitem[{Kahane et~al.(2017)Kahane, Courtin, and Gerdes}]{kahane-17}
Sylvain Kahane, Marine Courtin, and Kim Gerdes. 2017.
\newblock \href {https://www.aclweb.org/anthology/W17-7622} {Multi-word
  annotation in syntactic treebanks - {P}ropositions for {U}niversal
  {D}ependencies}.
\newblock In \emph{Proc. of the 16th International Workshop on Treebanks and
  Linguistic Theories}, pages 181--189, Prague, Czech Republic.

\bibitem[{Lin and Zeldes(2021)}]{LinZeldes2021}
Jessica Lin and Amir Zeldes. 2021.
\newblock \href {https://aclanthology.org/2021.law-1.18} {{W}iki{GUM}:
  Exhaustive entity linking for {W}ikification in 12 genres}.
\newblock In \emph{Proc. of The Joint 15th Linguistic Annotation Workshop (LAW)
  and 3rd Designing Meaning Representations (DMR) Workshop}, pages 170--175,
  Punta Cana, Dominican Republic.

\bibitem[{de~Marneffe et~al.(2021)de~Marneffe, Manning, Nivre, and
  Zeman}]{de_marneffe-21}
{Marie-Catherine} de~Marneffe, Christopher~D. Manning, Joakim Nivre, and Daniel
  Zeman. 2021.
\newblock \href {https://doi.org/10.1162/coli_a_00402} {Universal
  {D}ependencies}.
\newblock \emph{Computational Linguistics}, 47(2):255--308.

\bibitem[{Nivre et~al.(2016)Nivre, de~Marneffe, Ginter, Goldberg, Haji\v{c},
  Manning, {McDonald}, Petrov, Pyysalo, Silveira, Tsarfaty, and
  Zeman}]{nivre-16}
Joakim Nivre, {Marie-Catherine} de~Marneffe, Filip Ginter, Yoav Goldberg, Jan
  Haji\v{c}, Christopher~D. Manning, Ryan {McDonald}, Slav Petrov, Sampo
  Pyysalo, Natalia Silveira, Reut Tsarfaty, and Daniel Zeman. 2016.
\newblock \href
  {http://www.lrec-conf.org/proceedings/lrec2016/pdf/348_Paper.pdf} {Universal
  {D}ependencies v1: a multilingual treebank collection}.
\newblock In \emph{Proc. of {LREC}}, pages 1659--1666, Portoro\v{z}, Slovenia.

\bibitem[{Nivre et~al.(2020)Nivre, de~Marneffe, Ginter, Haji\v{c}, Manning,
  Pyysalo, Schuster, Tyers, and Zeman}]{nivre-20}
Joakim Nivre, {Marie-Catherine} de~Marneffe, Filip Ginter, Jan Haji\v{c},
  Christopher~D. Manning, Sampo Pyysalo, Sebastian Schuster, Francis Tyers, and
  Daniel Zeman. 2020.
\newblock \href {https://www.aclweb.org/anthology/2020.lrec-1.497} {Universal
  {D}ependencies v2: {A}n evergrowing multilingual treebank collection}.
\newblock In \emph{Proc. of {LREC}}, pages 4027--4036, Marseille, France.

\bibitem[{Peng and Zeldes(2018)}]{PengZeldes2018}
Siyao Peng and Amir Zeldes. 2018.
\newblock \href {https://www.aclweb.org/anthology/W18-4918} {All roads lead to
  {UD}: {C}onverting {S}tanford and {P}enn parses to {E}nglish {U}niversal
  {D}ependencies with multilayer annotations}.
\newblock In \emph{Proc. of the Joint Workshop on Linguistic Annotation,
  Multiword Expressions and Constructions ({LAW-MWE-CxG-2018})}, pages
  167--177, Santa Fe, New Mexico, {USA}.

\bibitem[{Poesio et~al.(2018)Poesio, Grishina, Kolhatkar, Moosavi, Roesiger,
  Roussel, Simonjetz, Uma, Uryupina, Yu, and Zinsmeister}]{PoesioEtAl2018}
Massimo Poesio, Yulia Grishina, Varada Kolhatkar, Nafise Moosavi, Ina Roesiger,
  Adam Roussel, Fabian Simonjetz, Alexandra Uma, Olga Uryupina, Juntao Yu, and
  Heike Zinsmeister. 2018.
\newblock \href {https://doi.org/10.18653/v1/W18-0702} {Anaphora resolution
  with the {ARRAU} corpus}.
\newblock In \emph{Proc. of the First Workshop on Computational Models of
  Reference, Anaphora and Coreference (CRAC 2018)}, pages 11--22, New Orleans,
  LA.

\bibitem[{Popel et~al.(2017)Popel, \v{Z}abokrtsk\'{y}, and
  Vojtek}]{PopelZabokrtskyVojtek2017}
Martin Popel, Zden\v{e}k \v{Z}abokrtsk\'{y}, and Martin Vojtek. 2017.
\newblock \href {https://aclanthology.org/W17-0412} {Udapi: Universal {API} for
  {U}niversal {D}ependencies}.
\newblock In \emph{Proc. of the {NoDaLiDa} 2017 Workshop on Universal
  Dependencies ({UDW} 2017)}, pages 96--101, Gothenburg, Sweden.

\bibitem[{Ratinov et~al.(2011)Ratinov, Roth, Downey, and
  Anderson}]{RatinovEtAl2011}
Lev Ratinov, Dan Roth, Doug Downey, and Mike Anderson. 2011.
\newblock \href {https://aclanthology.org/P11-1138} {Local and global
  algorithms for disambiguation to {W}ikipedia}.
\newblock In \emph{Proc. of ACL-HLT}, pages 1375--1384, Portland, Oregon, USA.
  Association for Computational Linguistics.

\bibitem[{Ruppenhofer et~al.(2016)Ruppenhofer, Ellsworth, Petruck, Johnson,
  Baker, and Scheffczyk}]{ruppenhofer-16}
Josef Ruppenhofer, Michael Ellsworth, Miriam R.~L. Petruck, Christopher~R.
  Johnson, Collin~F. Baker, and Jan Scheffczyk. 2016.
\newblock \href {https://framenet2.icsi.berkeley.edu/docs/r1.7/book.pdf}
  {{FrameNet} {II}: extended theory and practice}.

\bibitem[{Schneider and Zeldes(2021)}]{mncs-udw}
Nathan Schneider and Amir Zeldes. 2021.
\newblock \href {https://aclweb.org/anthology/2021.udw-1.14} {Mischievous
  nominal constructions in {U}niversal {D}ependencies}.
\newblock In \emph{Proc. of the Fifth Workshop on Universal Dependencies (UDW,
  SyntaxFest 2021)}, pages 160--172, Sofia, Bulgaria.

\bibitem[{Silveira et~al.(2014)Silveira, Dozat, de~Marneffe, Bowman, Connor,
  Bauer, and Manning}]{silveira-14}
Natalia Silveira, Timothy Dozat, {Marie-Catherine} de~Marneffe, Samuel~R.
  Bowman, Miriam Connor, John Bauer, and Christopher~D. Manning. 2014.
\newblock \href
  {http://www.lrec-conf.org/proceedings/lrec2014/pdf/1089_Paper.pdf} {A gold
  standard dependency corpus for {E}nglish}.
\newblock In \emph{Proc. of {LREC}}, pages 2897--2904, Reykjav\'{i}k, Iceland.

\bibitem[{Smith(1999)}]{smith-99}
Jeffrey~D. Smith. 1999.
\newblock English number names in {HPSG}.
\newblock In Gert Webelhuth, {Jean-Pierre} Koenig, and Andreas Kathol, editors,
  \emph{Lexical and Constructional Aspects of Linguistic Explanation}, pages
  145--160. {CSLI} Publications, Stanford, {CA}.

\bibitem[{Uryupina et~al.(2020)Uryupina, Artstein, Bristot, Cavicchio, Delogu,
  Rodriguez, and Poesio}]{UryupinaEtAl2019}
Olga Uryupina, Ron Artstein, Antonella Bristot, Federica Cavicchio, Francesca
  Delogu, Kepa~J. Rodriguez, and Massimo Poesio. 2020.
\newblock \href {https://doi.org/10.1017/S1351324919000056} {Annotating a broad
  range of anaphoric phenomena, in a variety of genres: the {ARRAU} {C}orpus}.
\newblock \emph{Natural Language Engineering}, 26:95--128.

\bibitem[{Zeldes(2017)}]{Zeldes2017b}
Amir Zeldes. 2017.
\newblock \href {https://doi.org/http://dx.doi.org/10.1007/s10579-016-9343-x}
  {The {GUM} corpus: Creating multilayer resources in the classroom}.
\newblock \emph{Language Resources and Evaluation}, 51(3):581--612.

\bibitem[{Zeldes and Abrams(2018)}]{ZeldesAbrams2018}
Amir Zeldes and Mitchell Abrams. 2018.
\newblock \href {https://www.aclweb.org/anthology/W18-6022} {The {C}optic
  {U}niversal {D}ependency {T}reebank}.
\newblock In \emph{Proc. of the Second Workshop on Universal Dependencies
  ({UDW} 2018)}, pages 192--201, Brussels, Belgium.

\bibitem[{Zeman(2021)}]{zeman-21}
Daniel Zeman. 2021.
\newblock \href {https://aclweb.org/anthology/2021.udw-1.15} {Date and time in
  {U}niversal {D}ependencies}.
\newblock In \emph{Proc. of the Fifth Workshop on Universal Dependencies (UDW,
  SyntaxFest 2021)}, pages 173--193, Sofia, Bulgaria.

\end{thebibliography}
\bibliographystyle{acl_natbib}

% \appendix

\end{document}